\documentclass[conference]{IEEEtran}
\IEEEoverridecommandlockouts
\usepackage{cite}
\usepackage{amsthm}
\theoremstyle{definition}

\usepackage{amsmath,amssymb,amsfonts}
\usepackage{threeparttable}
\usepackage{graphicx}
\usepackage{textcomp}
\usepackage{algorithm}
\usepackage{algpseudocode}
\usepackage{enumitem}   
\usepackage{subfigure}

\usepackage[dvipsnames]{xcolor}
\definecolor{darkyellow}{rgb}{0.85, 0.65, 0.13} 
\usepackage{hyperref}

\begin{document}

\title{
Adapting GT2-FLS for Uncertainty Quantification: A Blueprint Calibration Strategy\\
\thanks{This work was supported by MathWorks\textsuperscript{\textregistered} in part by a Research Grant awarded to T. Kumbasar. Any opinions, findings, conclusions, or recommendations expressed in this paper are those of the authors and do not necessarily reflect the views of MathWorks, Inc.}

}


 \author{\IEEEauthorblockN{Yusuf Güven}
 \IEEEauthorblockA{\textit{Artificial Intelligence and Intelligent Systems Lab.} \\
 \textit{Istanbul Technical University}\\
 Istanbul, Türkiye \\
 guveny18@itu.edu.tr}
 \and
 \IEEEauthorblockN{Tufan Kumbasar}
 \IEEEauthorblockA{\textit{Artificial Intelligence and Intelligent Systems Lab.} \\
 \textit{Istanbul Technical University}\\
 Istanbul, Türkiye \\
 kumbasart@itu.edu.tr}}

\maketitle

\begin{abstract}
Uncertainty Quantification (UQ) is crucial for deploying reliable Deep Learning (DL) models in high-stakes applications. Recently, General Type-2 Fuzzy Logic Systems (GT2-FLSs) have been proven to be effective for UQ, offering Prediction Intervals (PIs) to capture uncertainty. However, existing methods often struggle with computational efficiency and adaptability, as generating PIs for new coverage levels $(\phi_d)$ typically requires retraining the model. Moreover, methods that directly estimate the entire conditional distribution for UQ are computationally expensive, limiting their scalability in real-world scenarios.
This study addresses these challenges by proposing a blueprint calibration strategy for GT2-FLSs, enabling efficient adaptation to any desired $\phi_d$ without retraining. By exploring the relationship between $\alpha$-plane type reduced sets and uncertainty coverage, we develop two calibration methods: a lookup table-based approach and a derivative-free optimization algorithm. These methods allow GT2-FLSs to produce accurate and reliable PIs while significantly reducing computational overhead. Experimental results on high-dimensional datasets demonstrate that the calibrated GT2-FLS achieves superior performance in UQ, highlighting its potential for scalable and practical applications.
\end{abstract}


\section{Introduction}

\begin{figure}[htbp]
    \centering
    \includegraphics[width=0.50\textwidth]{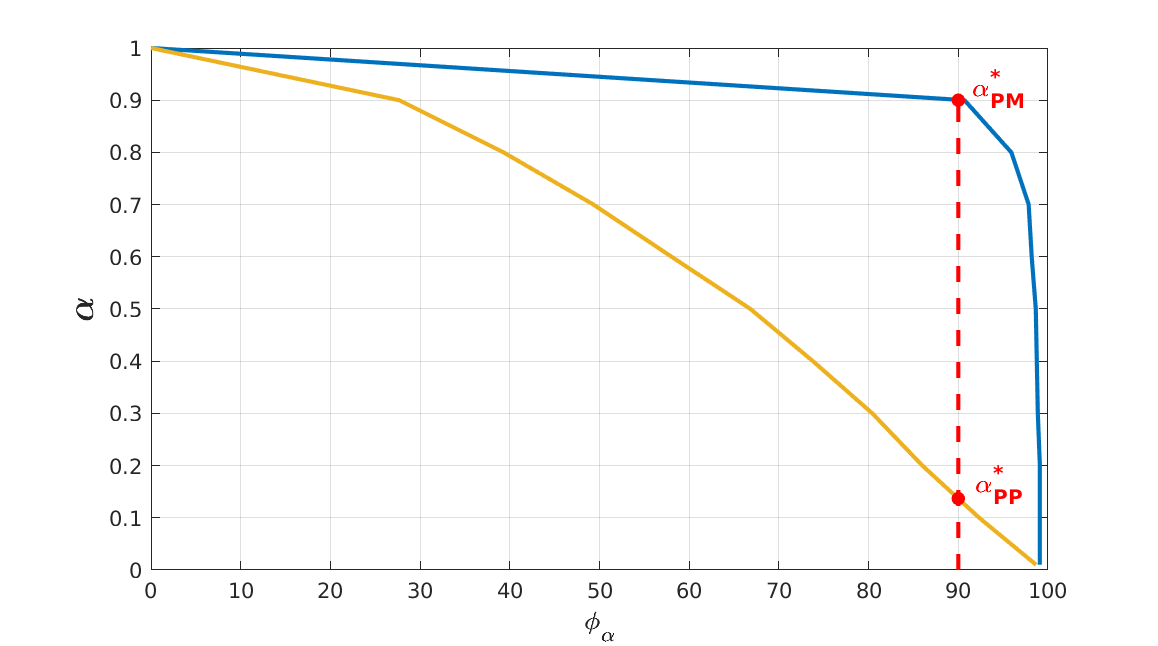}
    \caption{The {\color{blue}blue} and {\color{darkyellow} yellow} curves represent the calibration curves ($g^{-1}$) of the Parkinson's Motor (PM) and Powerplant (PP) datasets, respectively, where \({\color{red}\alpha_{\text{PM}}^*}\) and \({\color{red}\alpha_{\text{PP}}^*}\) indicate the critical \(\alpha^*\) achieving 90\% coverage, selected directly from the calibration curves without retraining the model. The calibration curves were obtained as follows: The baseline GT2-FLS was trained to generate PIs with $\phi_d=99\%$ through the TRS of its \(\alpha_0\) plane\([\underline{y}(\boldsymbol{x}, {\alpha_0}), \overline{y}({\boldsymbol{x}, \alpha_0})]\). After training, \(\alpha\)-planes were quantized as \([0.01, 0.1, 0.2, \dots, 1]\). For each quantized \(\alpha\)-plane, the bounds \([\underline{y}(\boldsymbol{x},{\alpha}), \overline{y}(\boldsymbol{x},{\alpha})]\) were obtained, and the correspond empiric coverage $(\phi_{\alpha})$ was calculated on the calibration dataset. Linear interpolation was applied via the \texttt{interp1} function to construct smooth calibration curves.}
    \label{fig:calibration_curve}
\end{figure}

Deep Learning (DL) increasingly impacts our lives by transforming industries, improving decision-making, and enabling smarter technologies across various domains. However, without reliability, safety, and consistency in real-world environments, the full potential of DL models remains unfulfilled \cite{psaros2023uncertainty}. Uncertainty Quantification (UQ) is essential in addressing these challenges, particularly in high-risk applications \cite{pearce, abdar2021review, 9612011}. 

For UQ, a promising model structure are the Type-2 Fuzzy Logic Systems which utilize Membership Functions (MFs) defined by interval type-2 or General Type-2 (GT2) Fuzzy Sets (FSs) \cite{sakalli, almaraash2023life, pekaslan2019leveraging, mendel2018comparing}. FLSs are widely used in applications requiring high accuracy, such as prediction and control systems \cite{shihabudheen2018recent, zheng2021fusion, wiktorowicz2023t2rfis, tavoosi2021review, han2021type}. Furthermore, some methods enhance the capabilities of T2-FLSs by generating Prediction Intervals (PIs) in addition to point-wise predictions, thereby improving the reliability and confidence of the model outputs\cite{beke, avci, ata}. Despite these advancements, these models may struggle to capture complex data characteristics such as non-Gaussian, skewed, asymmetric, and heteroscedastic aleatoric noise. To overcome this challenge, recent approaches have focused on directly estimating the entire conditional distribution through alternative loss functions \cite{SQR, Beyond_pinball, guven}. These methods enable the selection of a quantile-level pair to generate PIs with a desired coverage level $(\phi_d)$. Yet, learning the entire conditional distribution can be computationally expensive.




In this study, motivated by the aforementioned drawbacks, we raise the following research question \textit{"How can we adapt/calibrate a GT2-FLS trained for one coverage level $(\phi_d)$ to generate PIs for any other $\phi_d$ without retraining?"}. To address this, we start by exploring the connection between the Type Reduced Set (TRS) of IT2-FLS associated with an $\alpha$-plane $\alpha_k$ ($\alpha_k$-IT2-FLS) and the coverage of uncertainty. Based on our analysis, we propose a blueprint calibration strategy for GT2-FLSs to adapt trained GT2-FLS for any $\phi_d$, resulting in a Calibrated GT2-FLS (C-GT2-FLS).  

To develop the calibration methods for GT2-FLS, we first needed to answer: \textit{"Can we bridge the TRS of $\alpha_k$-IT2-FLS, extracted from a trained GT2-FLS, with its corresponding coverage level $\phi_{\alpha}$ ?"}. In other words, a mapping $g: \alpha \rightarrow \phi_{\alpha}$. Yet, for calibration, we required the inverse mapping $g^{-1}:\phi_{\alpha}  \rightarrow \alpha$. Due to the definition of coverage, it was not possible to provide a closed-form representation. To overcome this, we estimated the $\phi_{\alpha}$ through a calibration dataset by quantizing $\alpha$. Subsequently, we define a look-up table for representing $g^{-1}$ and visualize the calibration curve as shown in Fig.\ref{fig:calibration_curve}. This mapping offers a naive approach for selecting the $\alpha^*$-plane corresponding to any given $\phi_d$, enabling the calibration of the GT2-FLS. However, selecting an appropriate quantization level and determining an interpolation technique for the calibration curves introduces additional hyperparameters and design complexity. To address these challenges, instead of explicitly representing $g^{-1}$, we reformulated the calibration approach as a univariate optimization problem. We propose a derivative-free search algorithm over $\alpha$ to minimize the difference between $\phi_\alpha$ and the given $\phi_d$. 

To show the effectiveness of the proposed calibration framework, we compare the performance of C-GT2-FLS on high-dimensional datasets, against GT2-FLSs directly trained for \( \phi_d \). The results show that the calibration method over \( \alpha \)-planes effectively adapts the GT2-FLSs to produce accurate PI for given \( \phi_d \), without the need for retraining the GT2-FLS.





\section{Learning GT2-FLSs for UQ} \label{GT2-FLS}

This section briefly introduces the Zadeh-type GT2-FLS with its LPs and the dual-focused DL framework \cite{guven}.

\subsection{Inference of Zadeh-type GT2-FLSs}
\begin{figure}[t]
    \centering
    \includegraphics[width=0.45\textwidth]{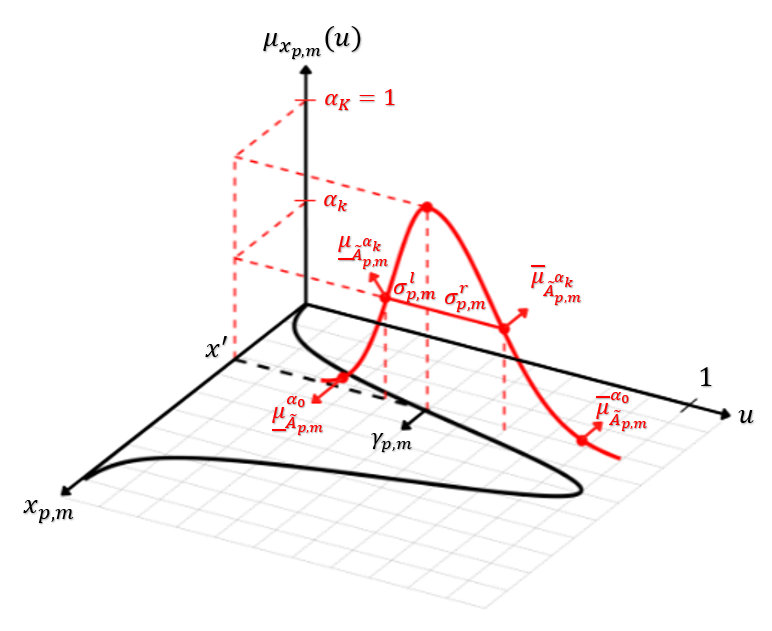} 
    \caption{Illustrations of a Z-GT2-FS with an $\alpha$ - plane}
    \label{Z-GT2-FS}
\end{figure}

The GT2-FLS is formulated for an input vector $\mathbf{x}=$ $\left(x_{1}, x_{2}, \ldots, x_{M}\right)^{\mathrm{T}}$  and a single output $y$. The rule base is composed of $P$ rules $(p=1,2, \ldots, P)$ that is defined as:
\begin{equation}\label{rule}
R_{p}: \text{If } x_{1} \text{ is } \tilde{A}_{p, 1} \text{ and} \ldots x_{M} \text{ is }\tilde{A}_{p, M} \text{ Then } y \text{ is } y_p 
\end{equation}
where $y_{p}$ represents the consequent MFs that are defined as:
\begin{equation}
    y_{p} = \sum_{m=1}^{M} a_{p, m} x_{m} + a_{p, 0}
    \label{eq1}
\end{equation}
The antecedent MFs are defined with GT2-FSs $\tilde{A}_{p, m}$ that are described through a collection of $\alpha$-planes $\left(\alpha_{k}\right)$ as follows:
\begin{equation}
    \tilde{A}_{p, m}=\bigcup_{\alpha_{k} \in[0,1]} \tilde{A}_{p, m}^{\alpha_{k}}
    \label{eq2}
\end{equation}
where $\tilde{A}_{p, m}^{\alpha_{k}}$ is the $\alpha$-plane of $\tilde{A}_{p, m}$ associated with $\alpha_{k} \in[0,1]$. 
In this study, we utilize the Zadeh representation of GT2-FS \cite{guven}. As illustrated in Fig. \ref{Z-GT2-FS}, the PMF is represented using a Type-1 FS $A_{p, m}$ defined as follows:
\begin{equation}
    {\mu}_{{A}_{p, m}}(x_m) = \exp \left (-\left (x_{m} - c_{p, m}\right)^{2} / 2\sigma_{p, m}^{2}\right )
    \label{pmf}
\end{equation}
We define an UMF and LMF of $\tilde{A}_{p, m}^{\alpha_{i}} (i \neq 0)$ as follows: 
\begin{equation} \label{GT2_mu_lower and_mu_upper}
\begin{split}
\overline{\mu}_{\tilde{A}_{p, m}^{\alpha_{k}}}(x_m)=\gamma_{{p, m}}(x_m)+\sqrt{-2 \ln \left(\alpha_k\right)} \sigma^r_{p, m} \\
\underline{\mu}_{\tilde{A}_{p, m}^{\alpha_{k}}}(x_m)=\gamma_{{p, m}}(x_m)-\sqrt{-2 \ln \left(\alpha_k\right)} \sigma^l_{p, m}  
\end{split}
\end{equation}
where $\sigma^l_{p, m}$ and $\sigma^r_{p, m}$ are the left and right standard deviations that define the shape and support of the SMF. $\gamma_{{p, m}}(x_m)$ is set by the PMF as 
$\gamma_{{p, m}}(x_m)={\mu}_{{A}_{p, m}}(x_m)$ via \eqref{pmf}. Note that we associate the $\alpha_{0}$-plane with $\alpha_0 \triangleq 0.01$ due to the domain space of $\ln(\cdot)$, which spans $(0, \infty]$ \cite{guven}. 

The output of GT2-FLS is as follows:
\begin{equation} \label{alphap}  y(\boldsymbol{x})=\frac{\sum_{k=0}^{K} y^{\alpha_{k}}(\boldsymbol{x}) \alpha_{k}}{\sum_{k=0}^{K} \alpha_{k}}
\end{equation}
where $\mathrm{y}^{\alpha_{k}}(\boldsymbol{x})$ is the output of an IT2-FLS associated with an $\alpha$-plane $\alpha_{k}\left(\alpha_{k}\right.$-IT2-FLS) that is defined as:
\begin{equation}
    y^{\alpha_{k}}(\boldsymbol{x})=
(\underline{y}^{\alpha_{k}}(\boldsymbol{x})+\overline{y}^{\alpha_{k}}(\boldsymbol{x}))/2
    \label{eq4}
\end{equation}
Here, $[\underline{y}^{\alpha_{k}}, \overline{y}^{\alpha_{k}}]$ is TRS of $\alpha_{k}$-IT2-FLS: 
\begin{equation}
\begin{split} 
\underline{y}^{\alpha_{k}}(\boldsymbol{x}) = \frac{\sum_{p=1}^{L} \underline{f}_{p}^{\alpha_{k}}(\boldsymbol{x}) {y}_{{p}} + \sum_{p=L+1}^{P} \overline{f}_{p}^{\alpha_{k}}(\boldsymbol{x}) {y}_{{p}}}{\sum_{p=1}^{L} \underline{f}_{p}^{\alpha_{k}}(\boldsymbol{x}) + \sum_{p=L+1}^{P} \overline{f}_{p}^{\alpha_{k}}(\boldsymbol{x})} \\
\overline{y}^{\alpha_{k}}(\boldsymbol{x}) = \frac{\sum_{p=1}^{R} \underline{f}_{p}^{\alpha_{k}}(\boldsymbol{x}) {y}_{{p}} + \sum_{p=R+1}^{P} \overline{f}_{p}^{\alpha_{k}}(\boldsymbol{x}) {y}_{{p}}}{\sum_{p=1}^{R} \underline{f}_{p}^{\alpha_{k}}(\boldsymbol{x}) + \sum_{p=R+1}^{P} \overline{f}_{p}^{\alpha_{k}}(\boldsymbol{x})}
\label{inference}
\end{split}
\end{equation}
where $L, R$ are the switching points of the Karnik-Mendel algorithm \cite{mendel_kitap}.  $\underline{f}_{p}^{\alpha_{k}}(\boldsymbol{x})$ and $\overline{f}_{p}^{\alpha_{k}}(\boldsymbol{x})$ are the lower and upper rule firing of the $p^{th}$ rule and are defined as:
\begin{equation}
\begin{split} 
\underline{f}_{p}^{\alpha_{k}}(\boldsymbol{x}) = \underline{\mu}_{\tilde{A}_{p, 1}^{\alpha_{k}}}\left(x_{1}\right) \cap \underline{\mu}_{\tilde{A}_{p, 2}^{\alpha_{k}}}\left(x_{2}\right) \cap \ldots \cap \underline{\mu}_{\tilde{A}_{p, M}^{\alpha_{k}}}\left(x_{M}\right) \\
\overline{f}_{p}^{\alpha_{k}}(\boldsymbol{x}) = \overline{\mu}_{\tilde{A}_{p, 1}^{\alpha_{k}}}\left(x_{1}\right) \cap \overline{\mu}_{\tilde{A}_{p, 2}^{\alpha_{k}}}\left(x_{2}\right) \cap \ldots \cap \overline{\mu}_{\tilde{A}_{p, M}^{\alpha_{k}}}\left(x_{M}\right)
\label{upper_firing}
\end{split}
\end{equation}
Here, $\cap$ denotes the t-norm operator \cite{mendel_kitap}. To handle the resulting curse of the dimensionality problem of $\cap$, we implemented the solutions presented in \cite{guven, koklu}.

\subsection{Learnable Parameter Sets}

The LP set of the GT2-FLS \( \boldsymbol{\theta} \) comprises the antecedent MF \( \boldsymbol{\theta}_{A} \) and the consequent MF \( \boldsymbol{\theta}_{C} \) parameters. $\boldsymbol{\theta}_{\mathrm{A}}$ is defined as $\{\boldsymbol{\theta}_{\mathrm{AP}},\boldsymbol{\theta}_{\mathrm{AS}}\}$, where $\boldsymbol{\theta}_{\mathrm{AP}}=\{\boldsymbol{c}, \boldsymbol{\sigma}\}$ with $\boldsymbol{c} = (c_{1,1}, \ldots, c_{P,M})^{T} \in \mathbb{R}^{P \times M}$, $\boldsymbol{\sigma} = (\sigma_{1,1}, \ldots,\sigma_{P,M})^{T} \in \mathbb{R}^{P \times M}$, and $\boldsymbol{\theta}_{\mathrm{AS}}=\{ \boldsymbol{\sigma^l}, \boldsymbol{\sigma^r}\}$ with $\boldsymbol{\sigma}^{(l)}=(\sigma_{1}^{(l)}, \ldots, \sigma_{M}^{(l)})^{T} \in \mathbb{R}^{M \times 1}$, and $\boldsymbol{\sigma}^{r} = (\sigma_{1}^{(r)}, \ldots, \sigma_{M}^{(r)})^{T} \in \mathbb{R}^{M \times 1}$.
$\boldsymbol{\theta}_{C}$ is defined as $\boldsymbol{\theta}_{\mathrm{C}} = \{ \boldsymbol{a}, \boldsymbol{a}_{0} \}$, with $\boldsymbol{a} = (\boldsymbol{a}_{1,1}, \ldots, \boldsymbol{a}_{P,M})^{T} \in \mathbb{R}^{P \times M}$ and  $\boldsymbol{a}_{0} = (a_{1,0}, \ldots, a_{P,0})^{T} \in \mathbb{R}^{P \times 1}$ . Note that, we set $\sigma^l_{p, m}=\sigma^l_{m}$ and $\sigma^r_{p, m}=\sigma^r_{m}, \forall p$. To sum up, GT2-FLS involves $(2P + 2)M + P(M + 1)$ LPs.

\subsection{DL Framework} \label{Dual-Focused}

Here, we outline the DL framework for GT2-FLS to achieve accurate predictions and high-quality PIs \cite{guven}. {Algorithm~1}  details the training process for a dataset $\left\{\boldsymbol{x}_{n}, y_{n}\right \}_{n=1}^{N}$, where $\boldsymbol{x}_{\boldsymbol{n}}=\left(x_{n, 1}, \ldots, x_{n, M}\right)^{T}$.

As we aim to learn a dual-focused GT2-FLS, the following loss is defined to be minimized by a DL optimizer \cite{beke, guven}:  

\begin{equation}
\theta^* = \arg \min_{\theta} L=\frac{1}{N} \sum_{n=1}^{N} \left[ L_R\left(\epsilon_n\right) + \ell\left(x_n, y_n, \underline{\tau}, \overline{\tau}\right) \right] \label{loss1}
\end{equation}
where $\epsilon_n = y_n - y(x_n)$.
For the accuracy-focused part $L_{R}(\cdot)$, we use the following  empirical risk function:
\begin{equation}
    L_{R}(\epsilon_n)= \log (\cosh (\epsilon_n))
    \label{emprical_risk}
\end{equation}
Whereas for the uncertainty-focused part, $\ell(\cdot)$ is constructed via a pinball loss \cite{Beyond_pinball}
and define the following loss:
\begin{equation}
    \ell\left(x_n, y_n,\underline{\tau}, \overline{\tau}\right) = \underline{\ell}_{\tau}^{\alpha_{0}}\left(x_n, y_n,\underline{\tau}\right) + \overline{\ell}_{\tau}^{\alpha_{0}}\left(x_n, y_n,\overline{\tau}\right) \label{eq:tilted} 
\end{equation}
with
\begin{equation}\label{eq:tilted_lower} 
\underline{\ell}_\tau^{\alpha_0}= \max(\underline{\tau}(y_{n}-\underline{y}^{\alpha_0}({x}_{n})),(\underline{\tau}-1)(y_{n}-\underline{y}^{\alpha_0}({x}_{n})))
\end{equation}
\begin{equation}\label{eq:tilted_upper}
    \overline{\ell}_\tau^{\alpha_0}=\max(\overline{\tau}(y_{n}-\overline{y}^{\alpha_0}({x}_{n})),(\overline{\tau}-1)(y_{n}-\overline{y}^{\alpha_0}({x}_{n})))
\end{equation}
where lower $(\underline{\tau})$ and upper $(\overline{\tau})$ quantile levels are utilized to generate an envelope that captures the desired level of uncertainty ($\varphi_d=[\underline{\tau},\overline{\tau}]$). We use TRS of $\alpha_0$-plane, $[\underline{y}^{\alpha_0}(x_{n}), \overline{y}^{\alpha_0}(x_{n})]$ as our lower and upper bound predictions.

It is worth underlining that the training of FLSs is defined with a constraint optimization problem as highlighted in \cite{beke}. To enable the use of widely adopted DL optimizers, we reformulated the learning problem by applying the parameterization tricks for FLSs as described in \cite{guven, beke}.

\begin{algorithm}  
\caption{DL-based Dual-Focused GT2-FLS}
\begin{algorithmic}[1]
\label{alg:integration}
\State \textbf{Input:} $N$ training samples $(x_{n},y_{n})^{N}_{n=1}, \phi = [\underline{\tau}, \overline{\tau}]$
\State $K+1$, number of $\alpha$-planes 
\State $P$, number of rules
\State $mbs$, mini-batch size
\State \textbf{Output:} LP set ${\theta}$
\State Initialize ${\theta} = [{\theta}_{A}, {\theta}_{C}]$;
\For{\textbf{each } $mbs$ in $N$} 
    \State $\mu \leftarrow \text{PMF}(x; {{\theta}}_{{AP}})$ \Comment{ Eq. \eqref{pmf}} 
    \State $[\underline{\mu}^{\alpha_0}, \overline{\mu}^{\alpha_0}] \leftarrow \text{SMF}(\mu; {{\theta}}_{{AS}})$ 
    \Comment{ Eq. \eqref{GT2_mu_lower and_mu_upper}}
    \State $[\underline{y}^{\alpha_0}, \overline{y}^{\alpha_0}, y] \leftarrow \text{Inference}(\underline{\mu}^{\alpha_0}, \overline{\mu}^{\alpha_0}; {\theta}_C)$ \Comment{Eq. \eqref{inference}}
    \State Compute $L$ \Comment{Eq. \eqref{loss1}}
    \State Compute the gradient ${\partial L}/{\partial {\theta}}$
    \State Update ${\theta}$ via Adam optimizer
\EndFor

\State ${\theta}^* = {\arg \min }(L)$

\State \textbf{Return} $\theta^*$
\end{algorithmic}
\end{algorithm}

\section{Adapting GT2-FLS for UQ: The blueprint} \label{sec:calibration}

The DL framework presented in Section \ref{Dual-Focused} effectively captures the desired level of UQ for a given $\varphi_d=[\underline{\tau},\overline{\tau}]$ through the TRS of $\alpha_0$-IT2-FLS. Yet, if it is desired to quantify the uncertainty at different levels of $\varphi_d$, the GT2-FLS must be retrained. In the literature, several methods achieve this by learning all quantile levels in a single training session and selecting the desired ones to generate PIs \cite{SQR, Beyond_pinball, guven}.  Yet, these approaches incur significant computational costs. 

Motivated by this research challenge, we pose and answer the research question \textit{"How can we generate a PI for any desired coverage level $(\phi_d)$ without retraining a GT2-FLS from scratch?"}. To address this, we ask \textit{"Can we bridge the TRS of $\alpha_k$-IT2-FLS, extracted from a trained GT2-FLS ($\boldsymbol{\theta^*}$), with $\phi_{\alpha} $ generated by $[\underline{y}^{\alpha_{k}}(\boldsymbol{x}), \overline{y}^{\alpha_{k}}(\boldsymbol{x})]$ ?"}. Mathematically, we seek a function ($g$) such that: 
\begin{equation}  
g: \alpha \xrightarrow{\boldsymbol{\theta^*}} \phi_{\alpha}  
\end{equation}
In this context, we start by reformulating \eqref{inference} as:
\begin{equation}  
    [\underline{y}^{\alpha}(\boldsymbol{x}), \overline{y}^{\alpha}(\boldsymbol{x})] \xrightarrow{\boldsymbol{\theta^*}} [\underline{y}(\boldsymbol{x}, \alpha), \overline{y}(\boldsymbol{x}, \alpha)], \forall \alpha \in [0,1] 
\end{equation}
to transform $\alpha \in [0, 1]$ from a structural parameter of GT2-FLS to an input argument of GT2-FLS. This provides us extract $[\underline{y}(\boldsymbol{x}, \alpha), \overline{y}(\boldsymbol{x}, \alpha)]$ to calculate the coverage $\phi_{\alpha}, \forall \alpha \in [0,1]$. As the coverage calculation does not have a closed-form representation (unless the precise inverse cumulative distribution function is available), we can estimate $\phi_{\alpha}$ by calculating Prediction Interval Coverage Probability (PICP) via a left-out dataset (i.e. calibration dataset):
\begin{equation} \label{eq:picp}
PICP = \frac{1}{Q} \sum_{i=1}^Q \mathbb{I} \left( \underline{y}({x}_i, \alpha) \leq y_i \leq \overline{y}({x}_i, \alpha) \right)    
\end{equation}
With this formulation, the only thing we need to do is to \textbf{calibrate} $\alpha$ to find corresponding $[\underline{y}(\boldsymbol{x}, \alpha^*), \overline{y}(\boldsymbol{x}, \alpha^*)]$ which generates $\phi_d=\phi_{\alpha^*}$ without retraining the GT2-FLS. Thus, we need to represent the inverse of the function $g$: 
\begin{equation}
    g^{-1}:  \phi_d \xrightarrow{\boldsymbol{\theta^*}} \alpha
\end{equation}

Algorithm \ref{alg:overall} \footnote{MATLAB implementation. [Online]. Available: \url{https://tinyurl.com/2mr726xk}} presents the proposed learning framework, including a calibration phase. In the calibration step, we are slicing the trained GT2-FLS to find the best $\alpha^*$ that will result in $\phi_d$. Here, we emphasize that the GT2-FLS was trained with \(\phi = 99\%\) to capture a comprehensive range of uncertainties. Training at this high coverage level ensures that the GT2-FLS can effectively encompass the upper bounds of uncertainty, making it possible to derive PIs for other desired coverage levels \(\phi_d\). It is important to note that \(\phi_d\) must satisfy the condition \(\phi_d < 99\%\), as the training at \(\phi = 99\%\) establishes the maximum uncertainty envelope that the model can represent. This approach avoids retraining for each specific \(\phi_d\) and enables efficient generation of PIs across multiple levels. In the remaining part of the section, we present two calibration methods for obtaining a C-GT2-FLS.

\begin{algorithm}[H]
\caption{Adapting GT2-FLS for Calibrating UQ}
\label{alg:overall}
\textbf{Input:} Dataset, \(\phi_d\).\\
\textbf{Output:} Calibrated GT2-FLS$(\phi_d)$,  $\alpha^*$

\begin{algorithmic}[1]
    \State Partition the dataset into:
    \begin{itemize}
        \item Training set: \(\{(\boldsymbol{x}_n, y_n)\}_{n=1}^{N}\)
        \vspace{0.02cm}
        \item Calibration set: \((\boldsymbol{x}_q, y_q)_{q=1}^{Q}\) 
        \item Testing set: \((\boldsymbol{x}_m, y_m)_{m=1}^{M}\)
    \end{itemize}
    \State Train the GT2-FLS with \(\{(\boldsymbol{x}_n, y_n)\}_{n=1}^{N}\) via Algorithm 1 for \(\phi=99\%\).
    \State Calibrate the trained GT2-FLS (99\%) with \((\boldsymbol{x}_q, y_q)_{q=1}^{Q}\) to obtain \(\alpha^*\) satisfying \(\phi_d\).
    \State Test the C-GT2-FLS on \((\boldsymbol{x}_m, y_m)_{m=1}^{M}\).
\end{algorithmic}

\end{algorithm}

\subsection{Post-hoc Calibration Method-1: Look-up Table}

Here, we propose a naive calibration method by constructing a lookup table to represent \( g^{-1} \), allowing \( \alpha^* \) to be set for a given \( \phi_d \) by quantizing \( \alpha \) with a fixed step size \( \delta \). The calibration method is given in Algorithm \ref{alg:cal1}.  Although the method is easy to implement, it has the following disadvantages: 
\begin{itemize}
\item The choice of \( \delta \) introduces a trade-off between coverage quality and computational cost. Smaller values of \( \delta \) improve the precision of representing \( g^{-1} \) and better generate calibrated PIs, but this requires more function evaluations, increasing computational complexity.     
    \item The need to select an interpolation technique, which is a structural hyperparameter. The choice of method (e.g., linear, spline, or polynomial) affects both the accuracy and smoothness of the lookup table. Poor interpolation can lead to inaccuracies in representing \( g^{-1} \), leading to either underfitting or overfitting of the resulting PIs.     
\end{itemize}

\begin{algorithm}[h]
\caption{Post-hoc Calibration Method-1}
\label{alg:cal1}
\textbf{Input:} Calibration dataset, Quantization step size \(\delta\), \(\phi_d\).\\
\textbf{Output:} Optimal \(\alpha^*\).

\begin{algorithmic}[1]

    \State \textbf{Table Construction:} Quantize \(\alpha \in [0, 1]\) with step size \(\delta\):
    \[\alpha_{\delta} = [\alpha_0, \delta, 2\delta, \dots, 1] \in \mathbb{R}^K, \text{where } \alpha_0 = 0.01.\]

    \State For each \(\alpha \in \alpha_{\delta}\):
        \begin{itemize}
            \item Compute the TRS: 
            \[\underline{y}(\boldsymbol{x}, \alpha), \overline{y}(\boldsymbol{x}, \alpha).\]
            \item Calculate the PICP value using the calibration dataset to estimate \(\phi_{\alpha}\).
        \end{itemize}
        \item Construct a look-up table \(T\) with pairs \((\phi_{\alpha}, \alpha)\):
        \[T = [(\phi_{\alpha_0}, \alpha_0), (\phi_{\delta},\delta), \ldots, (\phi_1, 1)]\]
    
    \State \textbf{Interpolation Technique:} Define an interpolation method over \(T\) to represent the inverse mapping \(g^{-1}(\phi_{\alpha})\).

    \State \textbf{Inference:} Query \(T\) using \(\phi_d\) to obtain \(\alpha^*\).

\end{algorithmic}

\end{algorithm}

\subsection{Post-hoc Calibration Method-2: A search algorithm}
In this method, rather than trying to represent \( g^{-1} \), we transform the slicing procedure via the following optimization problem: 
\begin{equation}\label{a_opt}
\begin{split}
        \alpha^* = \arg\min_{\alpha} \|\phi_{\alpha} - \phi_d\|_1,  \quad \\ \text{s.t.} \quad \phi_{\alpha} = \text{PICP}([\underline{y}(\boldsymbol{x}, \alpha), \overline{y}(\boldsymbol{x}, \alpha)])
\end{split}
\end{equation}
where \( \phi_{\alpha} \) is the PICP computed via \([ \underline{y}(\boldsymbol{x}, \alpha), \overline{y}(\boldsymbol{x}, \alpha) ]\). Although \eqref{a_opt} is a simple univariate constrained optimization problem, the PICP constraint (defined in \eqref{eq:picp}) is not differentiable. Thus, we solved \eqref{a_opt} by developing a derivative-free search algorithm to find the optimal slice \( \alpha^* \). 

The GT2-FLS calibration method is summarized in Algorithm \ref{alg:cal2}. The algorithm starts with an initial value for \( \alpha^*\), computes the PICP, and iteratively adjusts \( \alpha^* \) by calculating the following two directions:  
\begin{equation}
\begin{aligned}
    \alpha^+ &= \min(\alpha^*+ \delta, 1) \\
    \alpha^- &= \max(\alpha^*- \delta, \alpha_0)
\end{aligned}
\end{equation}
Here, we employ the $\min$ and $\max$ operators to enforce the constraint \(\alpha \in [0.01, 1]\). Then, the optimizer evaluates \(\phi_\alpha\) in both directions \((\phi^+, \phi^-)\) and compares the results to determine the direction that minimizes \eqref{a_opt}, as follows:
\begin{equation}
\alpha^* := \begin{cases}
    \begin{aligned}
        &\alpha^+, \|\phi^+ - \phi_{d}\|_1 < \|\phi - \phi_{d}\|_1 \\
        &\alpha^-,  \|\phi^- - \phi_{d}\|_1 < \|\phi - \phi_{d}\|_1
    \end{aligned}
 \end{cases}
 \label{smf}
\end{equation}
If both directions show improvement, the algorithm selects the one yielding the greatest reduction in the measure. On the other hand, if neither direction results in an improvement, the step size \( \delta \) is scaled down by a factor \( \gamma \) to refine the search. This iterative process continues until $\phi_\alpha$ converges to \( \phi_d \) within a specified tolerance at which the optimal value \( \alpha^* \) is determined.

\begin{algorithm}[h]
\caption{Post-hoc Calibration Method-2}
\label{alg:cal2}

\begin{algorithmic}[1]
\State \textbf{Input:} $Q$ calibration samples, $(x_{q},y_{q})^{Q}_{q=1}$ 
\State \textbf{Argument:} $\phi_{d}$, $\alpha_{init}$, $\delta$, $\gamma$, $\epsilon$
\State \textbf{Output:} $\alpha^*$

\State Compute $[\underline{\textbf{y}}(\boldsymbol{x}, \alpha^*), \overline{\textbf{y}}(\boldsymbol{x}, \alpha^*)]$
\State Compute $\phi = \text{PICP}(\textbf{y}, [\underline{\textbf{y}}(\boldsymbol{x}, \alpha^*), \overline{\textbf{y}}(\boldsymbol{x}, \alpha^*)])$

\While{ $\|\phi - \phi_{d}\|_1 \geq \epsilon$}
    \State Update $\alpha^+ = \min(\alpha^*+ \delta, 1)$
    \State Update $\alpha^- = \max(\alpha^*- \delta, \alpha_0)$
    \vspace{0.05cm} 
    \State Compute $[\underline{\textbf{y}}(\boldsymbol{x}, \alpha^+), \overline{\textbf{y}}(\boldsymbol{x}, \alpha^+)]$
    \State Compute $\phi^+ = \text{PICP}(\textbf{y}, [\underline{\textbf{y}}(\boldsymbol{x}, \alpha^+), \overline{\textbf{y}}(\boldsymbol{x}, \alpha^+)])$
    \vspace{0.05cm} 
    \State Compute $[\underline{\textbf{y}}(\boldsymbol{x}, \alpha^-), \overline{\textbf{y}}(\boldsymbol{x}, \alpha^-)]$
    \vspace{0.05cm} 
    \State Compute $\phi^- = \text{PICP}(\textbf{y}, [\underline{\textbf{y}}(\boldsymbol{x}, \alpha^-), \overline{\textbf{y}}(\boldsymbol{x}, \alpha^-)])$
    \vspace{0.1cm} 
    \If{$\|\phi^+ - \phi_{d}\|_1 < \|\phi - \phi_{d}\|_1$ \textbf{and} $\|\phi^- - \phi_{d}\|_1 < \|\phi - \phi_{d}\|_1$}
        \If{$\|\phi^+ - \phi_{d}\|_1 < \|\phi^- - \phi_{d}\|_1$}
            \State $\alpha^* := \alpha^+$, $\phi := \phi^+$
        \Else
            \State $\alpha^* := \alpha^-$, $\phi := \phi^-$
        \EndIf
    \ElsIf{$\|\phi^+ - \phi_{d}\|_1 < \|\phi - \phi_{d}\|_1$}
        \State $\alpha^* := \alpha^+$, $\phi := \phi^+$
    \ElsIf{$\|\phi^- - \phi_{d}\|_1 < \|\phi - \phi_{d}\|_1$}
        \State $\alpha^* := \alpha^-$, $\phi := \phi^-$
    \Else
        \State $\delta := \delta \gamma$
        \State \textbf{Continue}
    \EndIf
\EndWhile
\State \textbf{Return} $\alpha^*$
\end{algorithmic}
\end{algorithm}


\begin{figure}[t]
    \centering
    \includegraphics[width=0.47\textwidth]{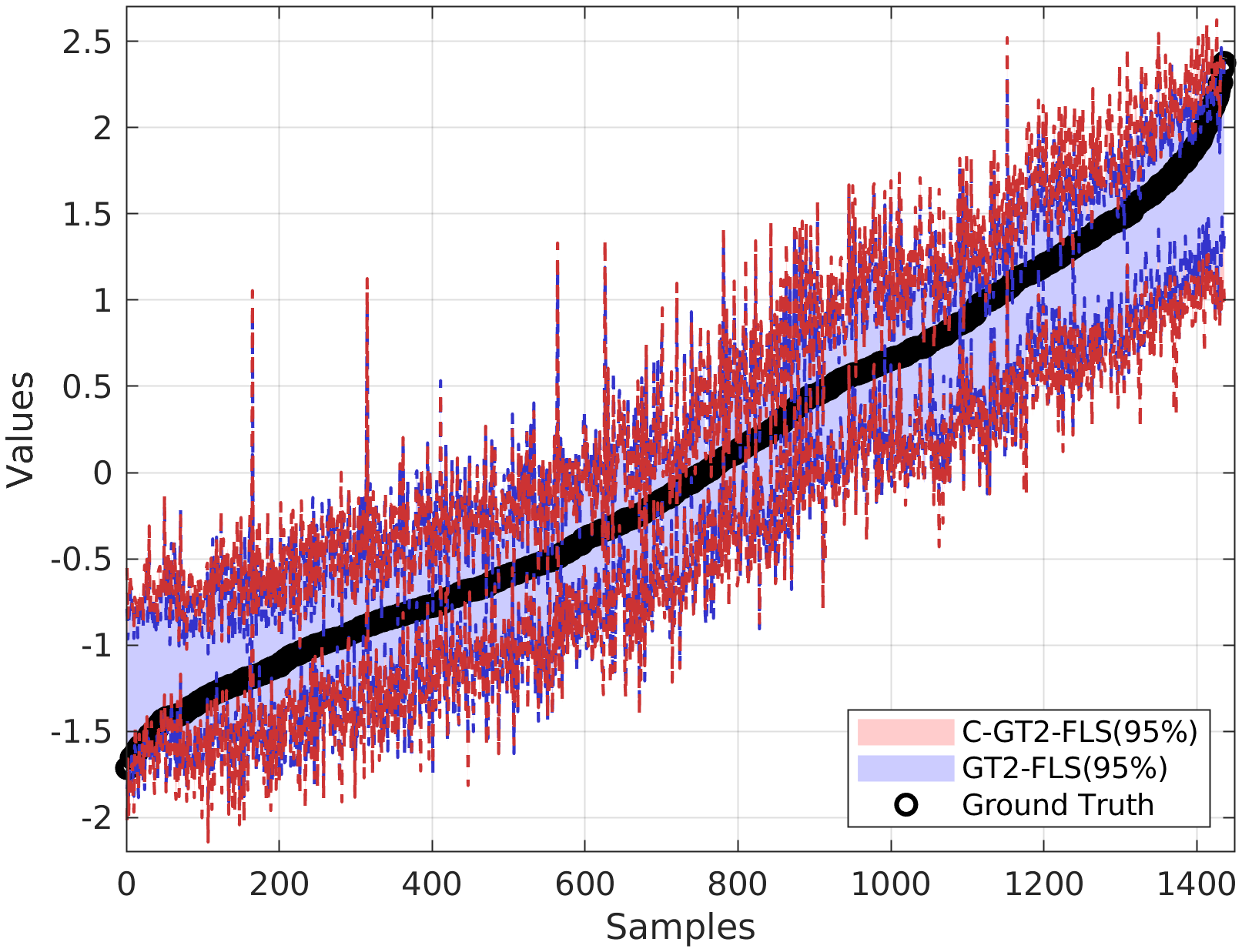} 
    \caption{Illustration of the PIs generated by GT2-FLS and C-GT2-FLS for the PP dataset: \textcolor{RubineRed}{C-GT2-FLS($95$\%)}: Calibrated for $\phi_d=95$ from a trained GT2-FLS ($99$\%); \textcolor{Blue}{GT2-FLS($95$\%)}: Trained GT2-FLS for $\phi_d=$95$\%$.}
    \label{fig:widevsnarrow}
\end{figure}

\section{Performance Analysis} \label{performance}

Here, we compare the performance of the C-GT2-FLS calibrated for $\phi_d$ against GT2-FLS trained for the same $\phi_d$. 

\subsection{Design of Experiments}
For evaluation, we utilize the following high-dimensional benchmark datasets: White Wine (WW), Parkinson's Motor UPDRS (PM), AIDS, and Powerplant (PP). The properties of these datasets are summarized in Table~\ref{tab:dataset-results-90}. All datasets were preprocessed using Z-score normalization.

We divided each dataset into training (70\%), calibration (15\%), and testing (15\%) sets. Initially, we trained a baseline GT2-FLS with $99\%$ coverage for all datasets. Using these trained GT2-FLS$(99\%)$, we extracted the following C-GT2-FLSs by applying Algorithm-3.

\begin{itemize}
    \item \textbf{C-GT2-FLS$(90\%)$:} We extract this model by slicing the trained GT2-FLS$(99\%)$ to achieve a coverage level of $\phi_d = 90\%$ by finding the best $\alpha$ plane, i.e. $\alpha^*$. 
    \item \textbf{C-GT2-FLS$(95\%)$:} We obtain the C-GT2-FLS$(95\%)$ by slicing the trained GT2-FLS$(99\%)$ to achieve a coverage level of $\phi_d = 95\%$ by determining $\alpha^*$.
\end{itemize}
To compare their coverage performances, we also trained the following GT2-FLSs specifically for $\phi_d\%$ using Algorithm-1: 
\begin{itemize}
    \item \textbf{GT2-FLS$(90 \%)$:} The GT2-FLS is trained by setting $\phi_d=90 \%$ ([$\underline{\tau},\overline{\tau}]=[0.05,0.95])$. 
    \item \textbf{GT2-FLS$(95 \%)$:} The GT2-FLS is trained by setting $\phi_d=95 \%$ ([$\underline{\tau},\overline{\tau}]=[0.025,0.975])$. 
\end{itemize}
For learning, each dataset was split into $85\%$ for training and $15\%$ for testing. The calibration data defined for C-GT2-FLS was incorporated into the training data of GT2-FLSs.

\subsection{Performance Evaluation} 
The experiments were conducted within MATLAB ${ }^{\circledR}$ and repeated with 5 different initial seeds for statistical analysis. We evaluated the performance using PICP and Prediction Interval Normalized Average Width (PINAW) \cite{PINAW}.

The mean performance metrics are presented in Table \ref{tab:dataset-results-90} and Table \ref{tab:dataset-results-95}. Note that, although we also report the RMSE and PINAW results, it is important to note that our C-GT2-FLSs are specifically calibrated to optimize PICP values for a given $\phi_d$. Therefore, the primary focus of our analysis lies in the PICP metric. We observe that:
\begin{itemize}
    \item For WW, the C-GT2-FLSs calibrated for 90\% and 95\% coverage levels deliver PICP values of 89.63\% and 94.66\%, respectively, effectively reaching the desired coverage levels without requiring training.  
    \item For PM, the C-GT2-FLSs achieved PICP values of 89.97\% at the 90\% coverage level and 94.42\% at the 95\% coverage level, showing better results compared to models trained directly for these coverage levels.
    \item For the AIDS dataset, the C-GT2-FLS achieves a calibrated PICP value of 94.40\% for $\phi_d=95\%$, showing improved performance over the model trained specifically for this coverage level. At the 90\% coverage level, the C-GT2-FLS achieves a calibrated PICP value of 87.42\%, outperforming the model trained for 90\%, emphasizing the effectiveness of the calibration. 
    \item For  PP, the C-GT2-FLSs achieve a calibrated PICP value of 95.02\% at the 95\% coverage level, surpassing the model trained directly for this coverage. At the 90\% coverage level, the C-GT2-FLS model achieves a calibrated PICP value of 87.86\%, again demonstrating better results than the model trained specifically for 90\%. These findings further emphasize the benefits of the search algorithm. in optimizing $\alpha$ to reach $\phi_d$ without training.
\end{itemize}

In summary, we conclude that the proposed calibration approach effectively transforms baseline GT2-FLSs into C-GT2-FLSs. This is evident as C-GT2-FLSs consistently outperform models trained directly for the same coverage rates in terms of PICP values. On the other hand, as illustrated in Fig.~\ref{fig:widevsnarrow} and from PINAW measures detailed in Tables~\ref{tab:dataset-results-90} and \ref{tab:dataset-results-95}, this approach results in wider PIs compared to GT2-FLSs that are directly optimized for $\phi_d$.

\begin{table}[t]
\centering
\caption{Performance Analysis Over 5 Experiments for $\phi_d=90\%$}
\begin{threeparttable}
\begin{tabular}{ccccc}
\hline
Dataset $(D \times N)$  & Metric & GT2-FLS$(90\%)$ & C-GT2-FLS$(90\%)$ \\
\hline
 & RMSE & \textbf{80.92($\pm$3.47)} & 81.42($\pm$4.52) \\
WW $(11 \times 4898)$    & PICP & 87.51($\pm$1.52) & \textbf{89.63($\pm$2.32)} \\
     & PINAW &  \textbf{41.80($\pm$4.77)} & 47.64($\pm$4.47) \\
\hline
  & RMSE & 60.50($\pm$4.65) & \textbf{59.67($\pm$4.31)} \\
 PM  $(19 \times 5875)$     & PICP & 91.51($\pm$1.84) & \textbf{89.97($\pm$1.74)}  \\
          & PINAW & \textbf{65.72($\pm$3.22)} & 74.11($\pm$5.93)  \\
\hline
& RMSE & {71.27($\pm$4.96)} & \textbf{70.86($\pm$2.63)}  \\
AIDS $(23 \times 2139)$      & PICP & 86.89($\pm$1.94) & \textbf{87.88($\pm$2.29)}  \\
     & PINAW & \textbf{89.81($\pm$14.02)} & 101.92($\pm$19.21)  \\
\hline
 & RMSE & \textbf{23.43($\pm$0.61)} & 23.57($\pm$0.62)  \\
   PP $(4 \times 9568)$  & PICP & 89.86($\pm$1.39) & \textbf{89.88($\pm$1.23)}  \\
     & PINAW & \textbf{17.32($\pm$0.73)} & 19.52($\pm$0.68)  \\
\hline
\end{tabular}
\begin{tablenotes}
\item (1) RMSE and PINAW values are scaled by 100. 
\item (2) Measures that are highlighted indicate the best ones.
\end{tablenotes}
\end{threeparttable}
\label{tab:dataset-results-90}
\end{table}

\section{Conclusion and Future Work} 
In this paper, we proposed the blueprint for solving the challenge of generating PIs for any desired coverage level $\phi_d$ in GT2-FLSs without retraining. By analyzing the relationship between $\alpha$-plane TRS and coverage rate, we developed a calibration framework with two approaches: a lookup table-based method and a derivative-free optimization algorithm. These methods enable efficient adaptation to varying $\phi_d$ levels, significantly improving computational efficiency and flexibility compared to existing methods.  
Our results demonstrate that the proposed C-GT2-FLS achieves comparable or superior performance to GT2-FLSs directly trained for specific coverage levels while eliminating the need for retraining.  

Future work will focus on developing calibration methods that ensure high-quality PIs by balancing width and coverage accuracy, further enhancing the framework's practicality for high-stakes applications.

\section*{Acknowledgment}
The authors acknowledge using ChatGPT to refine the grammar and enhance the English language expressions.

\bibliographystyle{IEEEtran}
\bibliography{IEEEabvr,cites}
\begin{table}[t]
\centering
\caption{Performance Analysis Over 5 Experiments for $\phi_d=95\%$}
\begin{threeparttable}
\begin{tabular}{ccccc}
\hline
Dataset $(D \times N)$ & Metric & GT2-FLS$(95\%)$ & C-GT2-FLS$(95\%)$ \\
\hline
 & RMSE & \textbf{81.15($\pm$3.48)} & 81.42($\pm$4.52)  \\
  WW $(11 \times 4898)$  & PICP & 92.93($\pm$0.89) & \textbf{94.66($\pm$1.60)}  \\
     & PINAW & \textbf{52.05($\pm$5.12)} & 60.67($\pm$6.69)  \\
\hline
 & RMSE &{60.59($\pm$4.03)} &  \textbf{59.67($\pm$4.31)}  \\
    PM $(19 \times 5875)$     & PICP & 96.34($\pm$1.55) & \textbf{94.42($\pm$0.97)}  \\
          & PINAW & \textbf{77.13($\pm$3.93)} & 88.83($\pm$3.92)  \\
\hline
 & RMSE & {71.31($\pm$3.76)} & \textbf{70.86($\pm$2.63)} \\
  AIDS $(23 \times 2139)$   & PICP & 92.29($\pm$1.41) & \textbf{94.40($\pm$2.20)}  \\
     & PINAW & \textbf{116.02($\pm$3.04)} & 146.41($\pm$9.89) \\
\hline
 & RMSE & \textbf{23.52($\pm$0.68)} & {23.57($\pm$0.62)} \\
  PP $(4 \times 9568)$  & PICP & 94.79($\pm$0.51) & \textbf{95.02($\pm$0.20)} \\
     & PINAW & \textbf{20.61($\pm$0.81)} & 23.09($\pm$1.00) \\
\hline
\end{tabular}
\begin{tablenotes}
\item (1) RMSE and PINAW values are scaled by 100. 
\item (2) Measures that are highlighted indicate the best ones.
\end{tablenotes}
\end{threeparttable}
\label{tab:dataset-results-95}
\end{table}
\end{document}